\documentclass[11pt]{article}

\usepackage[a4paper,margin=1in]{geometry}
\usepackage{amsmath,amssymb,mathtools,bm}
\usepackage{booktabs,multirow,array}
\usepackage{microtype}
\usepackage{xcolor}
\usepackage{graphicx}
\usepackage{subcaption}
\usepackage{enumitem}
\usepackage{natbib}
\usepackage{hyperref}
\usepackage[capitalise,noabbrev]{cleveref}
\usepackage{url}

\hypersetup{
  colorlinks=true,
  linkcolor=blue!55!black,
  citecolor=blue!55!black,
  urlcolor=blue!55!black,
  pdftitle={When the Preconditioning Exponent Turns Negative: Learning-Rate Coupling and Cross-Environment Generalization},
  pdfauthor={Gongyue Zhang and Honghai Liu}
}

\newcommand{\NumRuns}{420}
\newcommand{\NumCrossEvaluations}{1680}
\newcommand{\FinalEpochFraction}{92.38}
\newcommand{\WeakCrossGain}{0.35}
\newcommand{\ReversedCrossGain}{0.87}
\newcommand{\ReversedWorstGain}{2.48}
\newcommand{\WeakSpuriousRatioNegative}{0.381}
\newcommand{\WeakSpuriousRatioNonnegative}{0.435}
\newcommand{\WeakNoiseRatioNegative}{0.054}
\newcommand{\WeakNoiseRatioNonnegative}{0.066}
\newcommand{\ReversedSpuriousRatioNegative}{0.368}
\newcommand{\ReversedSpuriousRatioNonnegative}{0.447}

\title{When the Preconditioning Exponent Turns Negative:\
Learning-Rate Coupling and Cross-Environment Generalization}
\author{Gongyue Zhang and Honghai Liu}
\date{}

\begin{document}
\maketitle

\begin{abstract}
Adaptive optimizers are commonly parameterized by a fixed power of the second-moment estimate. Existing partially adaptive methods study exponents between momentum-like updates and the standard Adam square root, while the interaction between this exponent and the global learning rate is less understood. We perform a controlled cross-environment study using a paired four-environment classification problem with stable sparse features, environment-dependent spurious sparse features, dense features, and high-dimensional noise. Across \NumRuns{} source-training runs covering 21 preconditioning exponents $p\in[-0.5,0.5]$ and five learning rates $\eta\in[10^{-4},10^{-2}]$, we find that the exponent maximizing cross-environment accuracy decreases almost linearly with $\log_{10}\eta$. The fitted slopes range from $-0.270$ to $-0.300$, with $R^2$ between $0.972$ and $0.996$. At $\eta=10^{-2}$, source-validation selection still prefers positive exponents in all four environments, whereas cross-environment and worst-environment criteria prefer negative exponents. Checkpoint decomposition shows that lower $p$ reduces the learned spurious-to-stable and noise-to-stable weight ratios; under reversed correlation, it also reduces the magnitude of the harmful spurious margin. Negative $p$ is therefore not a universally optimal setting. It is a high-step-size allocation regime produced by the joint action of learning rate and preconditioning. The study also exposes a model-selection conflict: source-domain validation systematically selects a different preconditioning regime from the one that maximizes robustness to environmental change. The results are a single-seed, finite-budget mechanism study rather than a broad benchmark claim.
\end{abstract}

\section{Introduction}

Adaptive gradient methods change the geometry of optimization by assigning a different effective gain to each coordinate. Adam, for example, divides a bias-corrected first moment by the square root of a bias-corrected second moment \citep{kingma2015adam}. This coordinate-wise rescaling often accelerates optimization, but it can also select solutions that differ substantially from those reached by gradient descent \citep{wilson2017marginal}. A natural way to expose the strength of adaptivity is to replace the fixed square root with a continuous exponent,
\begin{equation}
\Delta\theta_{t,i}
= -\eta\,\widehat m_{t,i}(\widehat v_{t,i}+\epsilon)^{-p}.
\label{eq:update}
\end{equation}
The cases $p=0$ and $p=1/2$ recover a momentum-like direction and the usual Adam denominator, respectively. Partially adaptive methods have used $p\in(0,1/2]$ to interpolate between these behaviors \citep{chen2020padam}. More recent work has also questioned whether the square root is intrinsically necessary \citep{lin2024remove}.

Most discussions treat $p$ as an optimizer-family parameter and the learning rate $\eta$ as a separate scale parameter. This separation is exact only for a fixed update field. In multi-step training, $\eta$ changes the trajectory and therefore changes future gradients and moment estimates. Even before this feedback is considered, the instantaneous effective gain in \cref{eq:update} contains the product
\begin{equation}
\eta(\widehat v_{t,i}+\epsilon)^{-p}.
\end{equation}
The same nominal exponent can therefore correspond to very different effective coordinate gains at different learning rates. Conversely, a change in $p$ can partly compensate for a change in $\eta$.

This interaction is especially relevant under distribution shift. A model may fit both stable and spurious features on its source distribution, yet only the stable features remain predictive when the environment changes. Domain generalization and robust optimization usually address this problem through objectives, environment labels, or group information \citep{arjovsky2019irm,sagawa2019groupdro,gulrajani2021domainbed}. Here we ask a narrower question:

\begin{quote}
How does the optimizer's preconditioning exponent interact with learning rate to allocate a linear classifier across stable, spurious, and noise features, and where does the cross-environment optimum lie?
\end{quote}

We answer this question in a paired four-environment extension of the controlled construction used by \citet{wilson2017marginal}. Every environment shares the same labels, dense features, stable sparse features, sparse masks, amplitude noise, and pure-noise coordinates. Only the sign correlation of one sparse block with the label changes. A model is trained on one environment, its checkpoint is selected only by source validation loss, and the frozen checkpoint is evaluated on all four test environments. This design isolates optimizer-dependent feature allocation from unrelated dataset variation.

The central empirical result is a learning-rate-dependent reversal of the preferred exponent. For each source environment, the exponent maximizing mean cross-environment accuracy decreases nearly linearly with $\log_{10}\eta$. At small learning rates, the optimum remains positive. At large learning rates, the optimum enters the negative region. Source validation follows the same broad downward trend but remains systematically higher than the cross-environment optimum. Thus, negative $p$ should not be interpreted as a universally better optimizer. It is a regime that appears when a large global step is combined with a preconditioner that suppresses low-second-moment coordinates rather than amplifying them.

The contributions are:
\begin{enumerate}[leftmargin=1.5em]
  \item We conduct a complete $p\times\eta$ scan over four paired source environments, with target environments excluded from checkpoint selection.
  \item We identify a consistent empirical law: the cross-environment-optimal exponent decreases approximately linearly with $\log_{10}\eta$, with closely matched slopes across environments.
  \item We show that source-validation-optimal and cross-environment-optimal exponents are systematically separated, revealing an optimizer model-selection conflict under distribution shift.
  \item By decomposing the learned linear weights and signed margins, we show that lower and negative $p$ reduce relative reliance on spurious and noise coordinates in the high-learning-rate regime.
\end{enumerate}

The scope is deliberately limited. The study uses one random seed, one affine classifier, and a fixed training budget. Its purpose is to establish and explain a controlled phenomenon, not to claim a universal performance improvement on real-world domain generalization benchmarks.

\section{Related Work}

\paragraph{Adaptive optimization and implicit bias.}
AdaGrad and Adam use historical gradient magnitudes to construct coordinate-wise gains \citep{duchi2011adagrad,kingma2015adam}. \citet{wilson2017marginal} showed that adaptive methods can converge to solutions with different generalization properties from gradient descent, even on simple linearly separable problems. Padam introduced a continuous partial-adaptivity exponent between SGD-like and Adam-like updates and argued that excessive adaptivity can harm generalization \citep{chen2020padam}. \citet{lin2024remove} studied the opposite extension by removing the square root and strengthening the preconditioner. Our study differs in two ways. First, it scans the exponent symmetrically into the negative region. Second, it treats $p$ and the global learning rate as a coupled two-dimensional control space rather than evaluating $p$ at a fixed learning-rate regime.

\paragraph{Learning rate and feature learning.}
Large learning rates can change which patterns are learned and in what order. \citet{li2019largelr} connected large initial learning rates to delayed memorization of easy but poorly generalizing patterns. \citet{lu2024benign} showed that large-learning-rate oscillation can promote weak-feature learning in a controlled feature-noise model. These works motivate examining learning rate as more than a uniform multiplier. We complement them by showing that the exponent of adaptive preconditioning shifts systematically with learning rate, and that the robust regime can cross through zero into negative $p$.

\paragraph{Spurious correlations and domain generalization.}
Invariant risk minimization, group distributionally robust optimization, and DomainBed study learning under environmental or group shifts \citep{arjovsky2019irm,sagawa2019groupdro,gulrajani2021domainbed}. Recent controlled analyses emphasize that core and spurious features have distinct learning dynamics and that their relative complexity matters \citep{qiu2024complexity}. Our experiment does not propose a new domain generalization objective. Instead, it isolates how a standard source-only optimizer reallocates a fixed linear model across known stable, spurious, and noise blocks. This makes the optimizer mechanism directly observable.

\paragraph{Information allocation.}
The view that optimizer bias can be studied through relative signal allocation across parameter pathways has been developed in recent work \citep{zhang2026allocation}. The present paper applies the same general perspective to feature blocks under distribution shift: the preconditioning exponent changes which coordinates receive effective update mass, and this allocation has environment-dependent consequences.

\section{Learning-Rate--Exponent Coupling}

\subsection{Generalized preconditioning}

Let $g_{t,i}$ denote the stochastic gradient of coordinate $i$. We use Adam-style exponential moving averages
\begin{align}
 m_{t,i} &= \beta_1 m_{t-1,i} + (1-\beta_1)g_{t,i},\\
 v_{t,i} &= \beta_2 v_{t-1,i} + (1-\beta_2)g_{t,i}^2,
\end{align}
with standard bias corrections $\widehat m_{t,i}$ and $\widehat v_{t,i}$. The generalized update is given by \cref{eq:update}. For two coordinates $i$ and $j$, the instantaneous update ratio is
\begin{equation}
\frac{|\Delta\theta_{t,i}|}{|\Delta\theta_{t,j}|}
=
\frac{|\widehat m_{t,i}|}{|\widehat m_{t,j}|}
\left(
\frac{\widehat v_{t,j}+\epsilon}
{\widehat v_{t,i}+\epsilon}
\right)^p.
\label{eq:coordinate_ratio}
\end{equation}
The learning rate cancels from this same-step ratio, whereas $p$ directly changes coordinate allocation. Across steps, however, $\eta$ changes the parameters reached by the optimizer and therefore changes subsequent gradients, moments, and ratios.

For $p<0$, the preconditioner becomes
\begin{equation}
(\widehat v_{t,i}+\epsilon)^{-p}
=(\widehat v_{t,i}+\epsilon)^{|p|}.
\end{equation}
Coordinates with very small second moments are then suppressed rather than amplified. This does not imply that every high-$v$ coordinate is useful. It only reverses the usual Adam ordering of low- and high-second-moment coordinates.

\subsection{A local compensation relation}

To understand the joint trend, consider a representative coordinate scale $s=\widehat v+\epsilon$. The log effective gain is
\begin{equation}
\log G(\eta,p;s)=\log\eta-p\log s.
\end{equation}
If comparable training behavior requires this gain to remain near a constant $C$, then
\begin{equation}
 p \approx \frac{\log\eta-C}{\log s}.
\label{eq:compensation}
\end{equation}
When $0<s<1$, $\log s<0$, so
\begin{equation}
\frac{\partial p}{\partial\log\eta}
=\frac{1}{\log s}<0.
\label{eq:negative_slope}
\end{equation}
This simple relation predicts the direction observed in the experiments: increasing $\eta$ should reduce the exponent required to reach a comparable effective-update regime. It is not a full dynamical theory, because $s$ itself changes along the trajectory and differs across coordinates. It nevertheless supplies a testable sign prediction and explains why the optimum may pass below zero.

\subsection{Source and cross-environment optima}

For each source environment $e_s$, learning rate $\eta$, and exponent $p$, let $\theta^\star(e_s,\eta,p)$ be the checkpoint selected by source validation loss. We report three exponents:
\begin{align}
 p_{\mathrm{ID}}^\star(\eta)
 &= \arg\min_p L_{\mathrm{val}}^{e_s}(\theta^\star),\\
 p_{\mathrm{cross}}^\star(\eta)
 &= \arg\max_p \frac{1}{|\mathcal E|-1}
 \sum_{e\neq e_s} A_{\mathrm{test}}^e(\theta^\star),\\
 p_{\mathrm{worst}}^\star(\eta)
 &= \arg\max_p \min_{e\in\mathcal E} A_{\mathrm{test}}^e(\theta^\star).
\end{align}
The latter two are \emph{oracle analysis criteria}: target test environments are used only after source-based checkpoint selection. They are not deployable source-only model-selection rules. This distinction is essential because a major result of the paper is precisely that source selection and cross-environment selection disagree.

\section{Controlled Cross-Environment Experiment}

\subsection{Paired environments}

Labels $y\in\{0,1\}$ are sampled with $\Pr(y=1)=0.6$, and we write $\widetilde y=2y-1\in\{-1,+1\}$. Each input is a concatenation
\begin{equation}
 x=[x_{\mathrm{dense}},x_{\mathrm{stable}},x_{\mathrm{spur}},x_{\mathrm{noise}}].
\end{equation}
The four blocks are:
\begin{itemize}[leftmargin=1.5em]
  \item \textbf{Dense block:} four coordinates, each generated as $0.7\widetilde y+\mathcal N(0,3^2)$.
  \item \textbf{Stable sparse block:} 60 coordinates divided equally among activation probabilities $\pi\in\{0.005,0.01,0.02,0.05,0.1,0.2\}$. Active coordinates have signal magnitude $3.0$, additive noise standard deviation $0.05$, and label correlation $1.0$.
  \item \textbf{Spurious sparse block:} the same dimensions, activation probabilities, signal magnitude, and noise as the stable block, but with environment-dependent label correlation.
  \item \textbf{Noise block:} 3000 independent standard Gaussian coordinates.
\end{itemize}

The spurious correlations are $0.9$, $0.5$, $0$, and $-0.5$, defining strong-positive, weak-positive, neutral, and reversed environments. For a fixed split, all random draws are paired across environments. Labels, dense features, stable sparse features, sparse masks, amplitude noise, and pure-noise coordinates are identical. Only the sign correlation of the spurious block changes. This pairing makes cross-environment differences attributable to the intended shift rather than to resampling noise.

Each environment contains 8192 training, 2048 validation, and 8192 test examples. The input dimension is 3124.

\subsection{Model and optimization grid}

The model is a single affine binary classifier
\begin{equation}
 f_\theta(x)=x^\top w+b,
\end{equation}
trained with binary cross entropy. The optimizer uses \cref{eq:update} with $\beta_1=0.9$, $\beta_2=0.999$, $\epsilon=10^{-8}$, and zero weight decay. We scan
\begin{equation}
 p\in\{-0.50,-0.45,\ldots,0.45,0.50\}
\end{equation}
and
\begin{equation}
 \eta\in\{10^{-4},3\times10^{-4},10^{-3},3\times10^{-3},10^{-2}\}.
\end{equation}
For each of the four source environments, this gives $21\times5=105$ configurations and \NumRuns{} runs in total. Every run uses seed 42, batch size 128, and 50 epochs. The best checkpoint is chosen only by source validation loss. The selected checkpoint is then evaluated on all four test environments, yielding \NumCrossEvaluations{} cross-evaluations.

\subsection{Feature-allocation diagnostics}

Because the classifier is linear and the feature blocks are known, the learned allocation is directly measurable. For block $B$, we report
\begin{equation}
 R_B=\operatorname{RMS}(w_B),
\end{equation}
and especially
\begin{equation}
 R_{\mathrm{spur/stable}}
=\frac{\operatorname{RMS}(w_{\mathrm{spur}})}
{\operatorname{RMS}(w_{\mathrm{stable}})},
\qquad
R_{\mathrm{noise/stable}}
=\frac{\operatorname{RMS}(w_{\mathrm{noise}})}
{\operatorname{RMS}(w_{\mathrm{stable}})}.
\end{equation}
We also decompose the mean signed test margin by block:
\begin{equation}
M_B^e
=\mathbb E_{(x,y)\sim e}
[\widetilde y\,x_B^\top w_B].
\label{eq:margin}
\end{equation}
A negative $M_{\mathrm{spur}}^e$ means that the learned spurious component actively opposes the correct label in environment $e$.

\section{Results}

\subsection{The cross-environment-optimal exponent decreases with learning rate}

\begin{figure}[t]
  \centering
  \includegraphics[width=\linewidth]{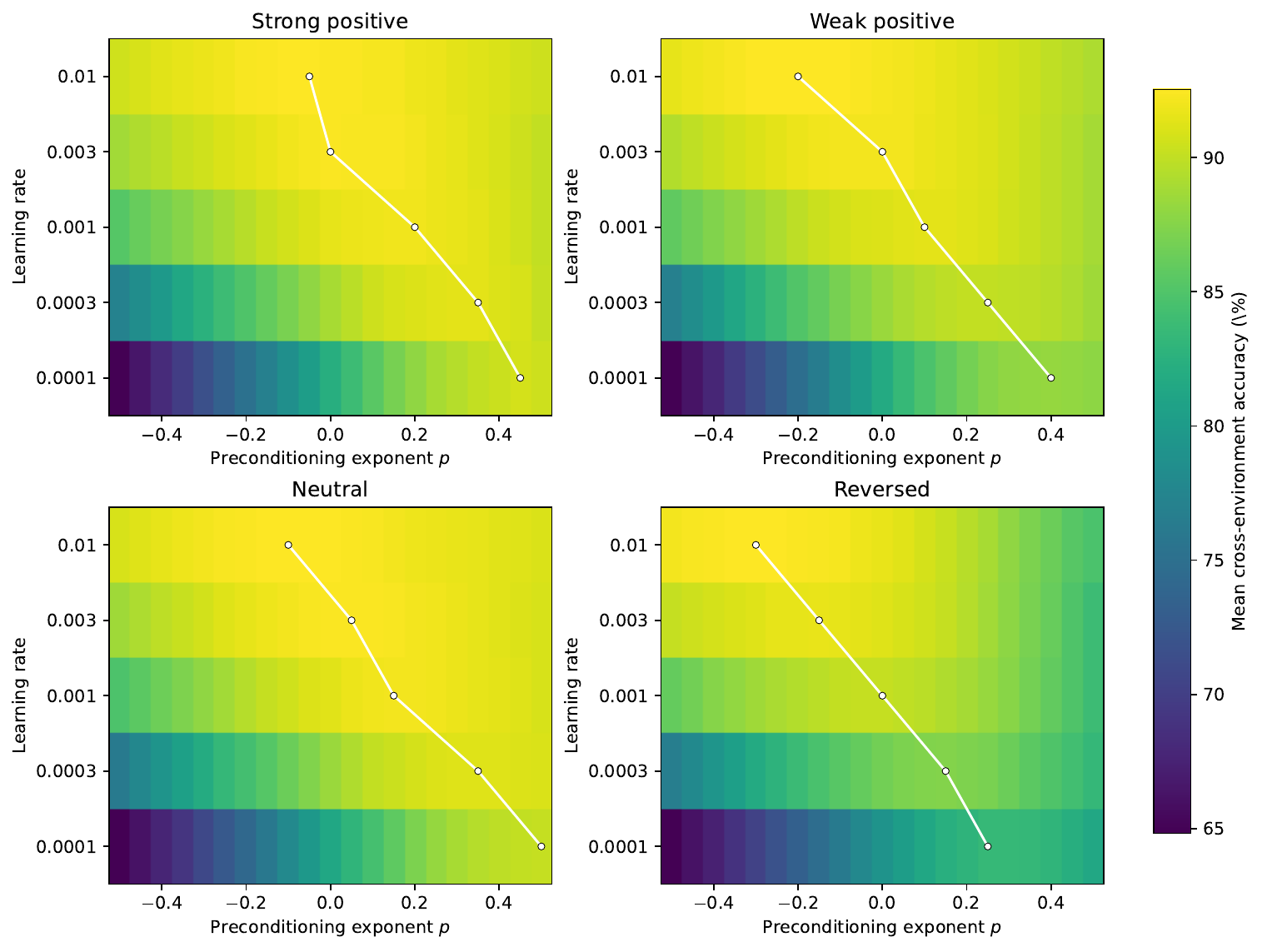}
  \caption{Mean cross-environment accuracy over the complete $p\times\eta$ grid. Each panel uses one source environment; the source test environment is excluded from the mean. White markers show the cross-environment-optimal exponent at each learning rate. The optimum moves monotonically toward lower $p$ as the learning rate increases.}
  \label{fig:heatmaps}
\end{figure}

\Cref{fig:heatmaps} shows mean cross-environment accuracy over the full $p\times\eta$ grid. The white path marks $p_{\mathrm{cross}}^\star$ at each learning rate. All four environments exhibit the same qualitative motion: the optimum shifts from the positive side at small $\eta$ toward zero and then into negative $p$ at large $\eta$.

Fitting
\begin{equation}
 p_{\mathrm{cross}}^\star=a\log_{10}\eta+b
\end{equation}
produces slopes between $-0.270$ and $-0.300$. The fit is unusually consistent across the four shifts, with $R^2$ values from $0.972$ to $0.996$ (\cref{tab:fits}). The result supports the sign prediction in \cref{eq:negative_slope}. It does not imply an exact universal linear law, but it does show that the observed reversal is structured rather than an isolated grid artifact.

\begin{table}[t]
  \centering
  \caption{Linear fits of $p_{\mathrm{cross}}^\star$ against $\log_{10}\eta$.}
  \label{tab:fits}
  \begin{tabular}{lrr}
\toprule
Source environment & Slope & $R^2$ \\
\midrule
Strong positive & -0.270 & 0.972 \\
Weak positive & -0.290 & 0.993 \\
Neutral & -0.300 & 0.992 \\
Reversed & -0.280 & 0.996 \\
\bottomrule
\end{tabular}
\end{table}

\begin{figure}[t]
  \centering
  \includegraphics[width=\linewidth]{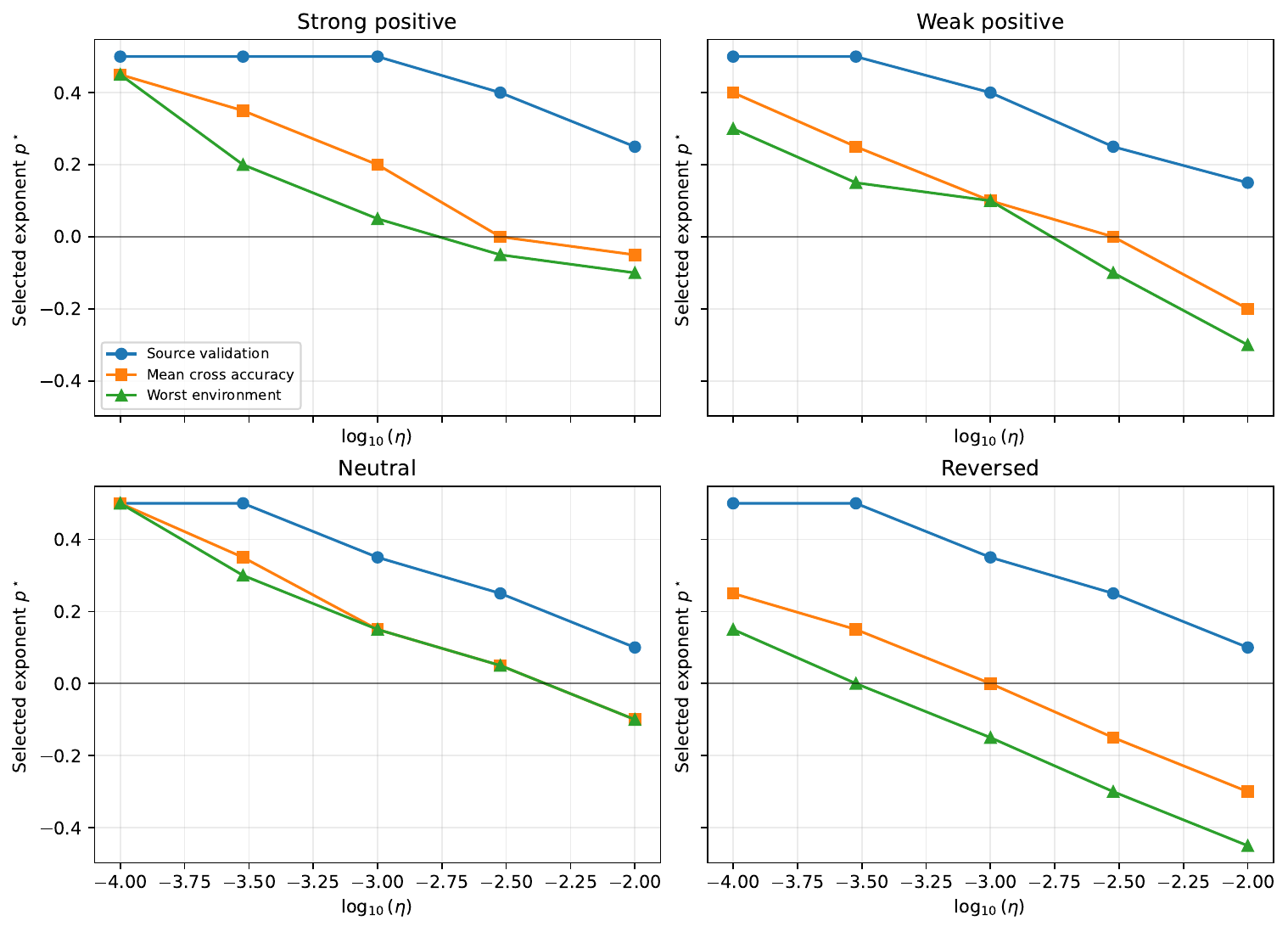}
  \caption{Exponent selected by source validation, mean cross-environment accuracy, and worst-environment accuracy. All three optima generally decline with learning rate, but the robust criteria are systematically lower and enter the negative region earlier.}
  \label{fig:optima}
\end{figure}

\subsection{Source validation and robust selection disagree}

\Cref{fig:optima} compares $p_{\mathrm{ID}}^\star$, $p_{\mathrm{cross}}^\star$, and $p_{\mathrm{worst}}^\star$. The source-validation optimum declines with learning rate but stays above the robust optima. At $\eta=10^{-2}$, source validation selects positive $p$ in every environment, while both robust criteria select negative $p$ (\cref{tab:highlr-optima}). The ID--cross gap ranges from $0.20$ to $0.40$.

\begin{table}[t]
  \centering
  \caption{Selected exponents at $\eta=10^{-2}$. Robust optima are evaluated only for analysis after source-validation checkpoint selection.}
  \label{tab:highlr-optima}
  \resizebox{\linewidth}{!}{\begin{tabular}{lrrrr}
\toprule
Source environment & $p_{\mathrm{ID}}^\star$ & $p_{\mathrm{cross}}^\star$ & $p_{\mathrm{worst}}^\star$ & ID--cross gap \\
\midrule
Strong positive & 0.25 & -0.05 & -0.10 & 0.30 \\
Weak positive & 0.15 & -0.20 & -0.30 & 0.35 \\
Neutral & 0.10 & -0.10 & -0.10 & 0.20 \\
Reversed & 0.10 & -0.30 & -0.45 & 0.40 \\
\bottomrule
\end{tabular}}
\end{table}

This gap has a direct methodological implication. Tuning the optimizer solely on source validation loss can favor a preconditioning regime that fits the source more aggressively but is less robust to a change in spurious correlation. Domain-generalization work has emphasized that model selection is part of the problem \citep{gulrajani2021domainbed}; here, even a one-dimensional optimizer exponent exhibits this conflict.

\subsection{Negative $p$ is a high-learning-rate regime, not a universal optimum}

\begin{figure}[t]
  \centering
  \includegraphics[width=\linewidth]{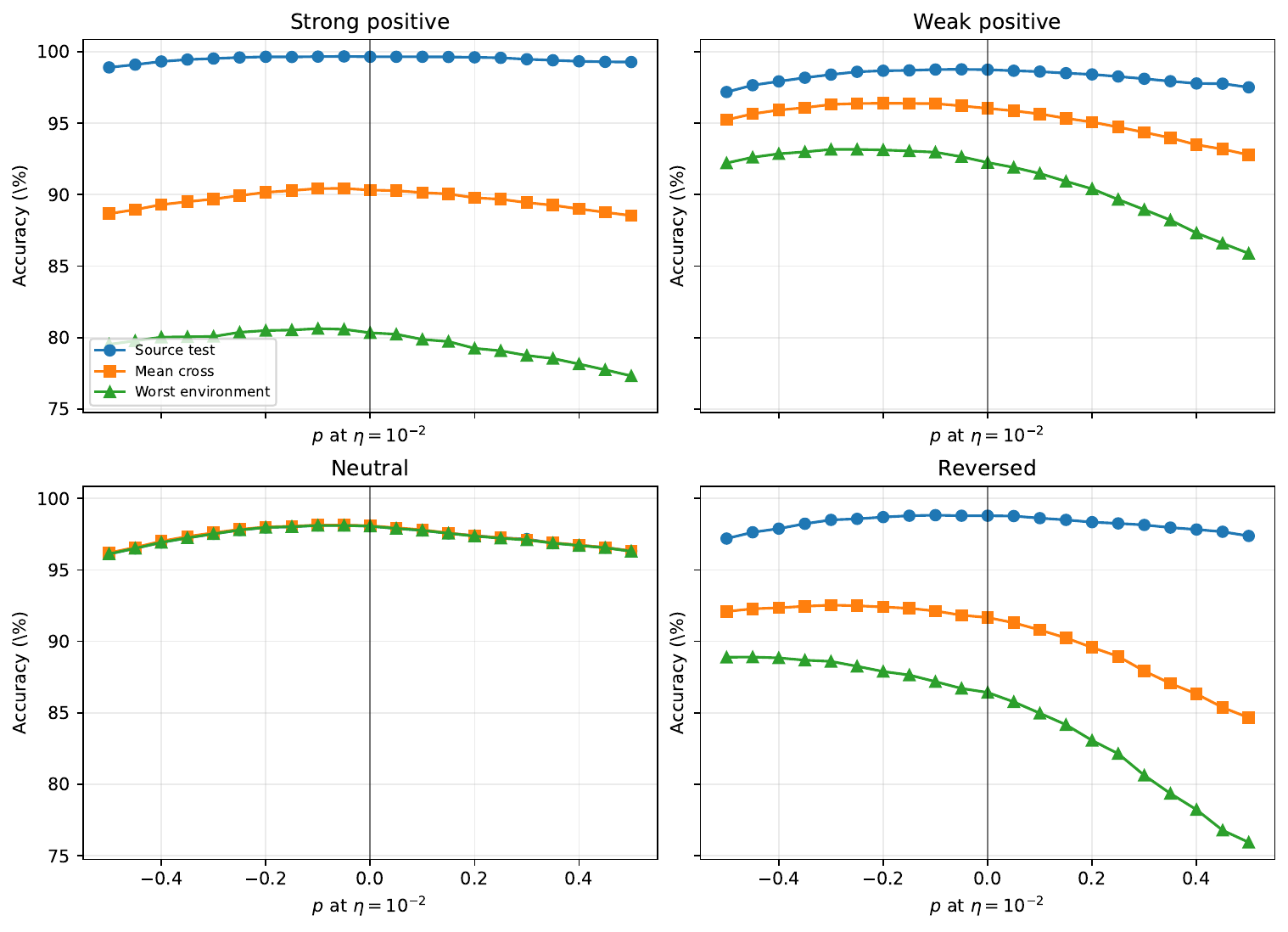}
  \caption{Source, mean cross-environment, and worst-environment accuracy at $\eta=10^{-2}$. Negative $p$ improves robustness relative to the best nonnegative setting in this high-step-size regime, but the optimum is finite and environment-dependent.}
  \label{fig:highlr-curves}
\end{figure}

\Cref{fig:highlr-curves} shows the full $p$ curves at $\eta=10^{-2}$. Negative $p$ does not uniformly improve every metric. Source accuracy is usually maximized at a higher exponent than cross or worst-environment accuracy. The robust curves peak at modestly negative values, while excessively negative $p$ eventually underfits useful structure. This produces a finite negative optimum rather than a monotonic preference for the smallest $p$.

Restricting the comparison to the best negative and best nonnegative exponent at $\eta=10^{-2}$, the cross-environment gains are small for strong-positive and neutral sources, but reach \WeakCrossGain{} percentage points for the weak-positive source and \ReversedCrossGain{} points for the reversed source (\cref{tab:negative}). The worst-environment gain for the reversed source is \ReversedWorstGain{} points. These are single-seed differences and should be read as mechanism magnitudes, not confidence intervals.

\begin{table}[t]
  \centering
  \caption{Best negative versus best nonnegative exponent for mean cross-environment accuracy at $\eta=10^{-2}$.}
  \label{tab:negative}
  \resizebox{\linewidth}{!}{\begin{tabular}{lrrrr}
\toprule
Source & Best negative $p$ & Cross acc. (\%) & Best $p\geq0$ & Gain (pp) \\
\midrule
Strong positive & -0.05 & 90.43 & 0.00 & +0.11 \\
Weak positive & -0.20 & 96.39 & 0.00 & +0.35 \\
Neutral & -0.10 & 98.14 & 0.00 & +0.05 \\
Reversed & -0.30 & 92.54 & 0.00 & +0.87 \\
\bottomrule
\end{tabular}}
\end{table}

\subsection{Lower $p$ reduces relative spurious and noise allocation}

\begin{figure}[t]
  \centering
  \includegraphics[width=\linewidth]{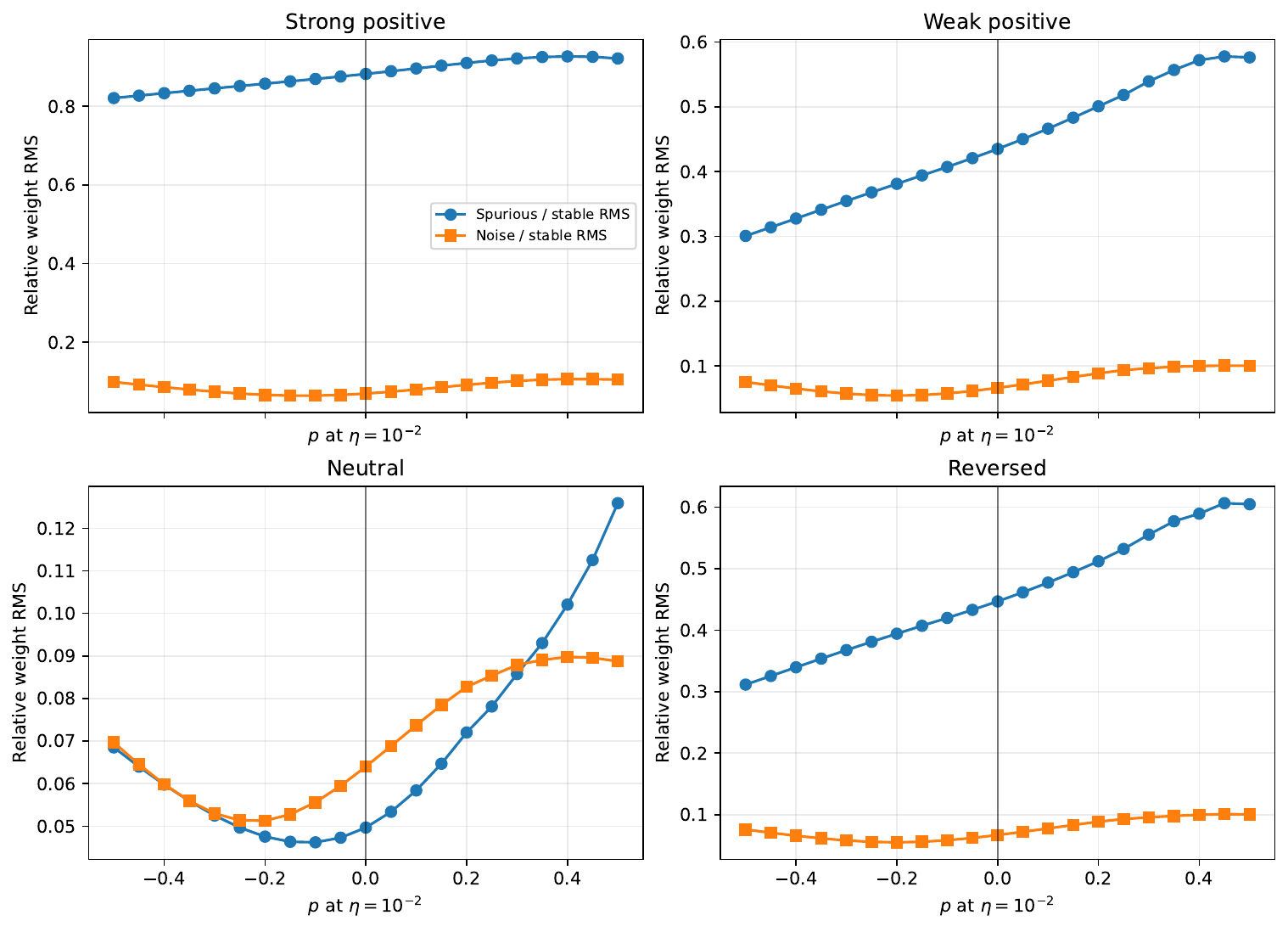}
  \caption{Learned feature allocation at $\eta=10^{-2}$. Lower $p$ generally reduces the RMS weight assigned to the spurious and pure-noise blocks relative to the stable sparse block. The neutral environment has little spurious signal and therefore shows a smaller, non-monotone ratio.}
  \label{fig:allocation}
\end{figure}

The checkpoint weights explain why the robust optimum moves left. \Cref{fig:allocation} plots the spurious-to-stable and noise-to-stable RMS ratios at $\eta=10^{-2}$. For the weak-positive and reversed sources, both ratios increase substantially with $p$. At the weak-positive cross optimum $p=-0.20$, the spurious-to-stable ratio is \WeakSpuriousRatioNegative{}, compared with \WeakSpuriousRatioNonnegative{} at the best nonnegative cross setting $p=0$. The noise-to-stable ratio falls from \WeakNoiseRatioNonnegative{} to \WeakNoiseRatioNegative{}. For the reversed source, the corresponding spurious-to-stable ratio falls from \ReversedSpuriousRatioNonnegative{} to \ReversedSpuriousRatioNegative{}.

The strong-positive source is a useful boundary case. Its spurious feature is highly predictive in the source and therefore remains heavily weighted across the grid. Negative $p$ yields only a small cross-environment improvement there. This confirms that the effect is not a generic shrinkage of all spurious coordinates. It depends on the competition between stable evidence, source correlation, training budget, and optimizer scaling.

\subsection{Margin decomposition under correlation reversal}

\begin{figure}[t]
  \centering
  \includegraphics[width=\linewidth]{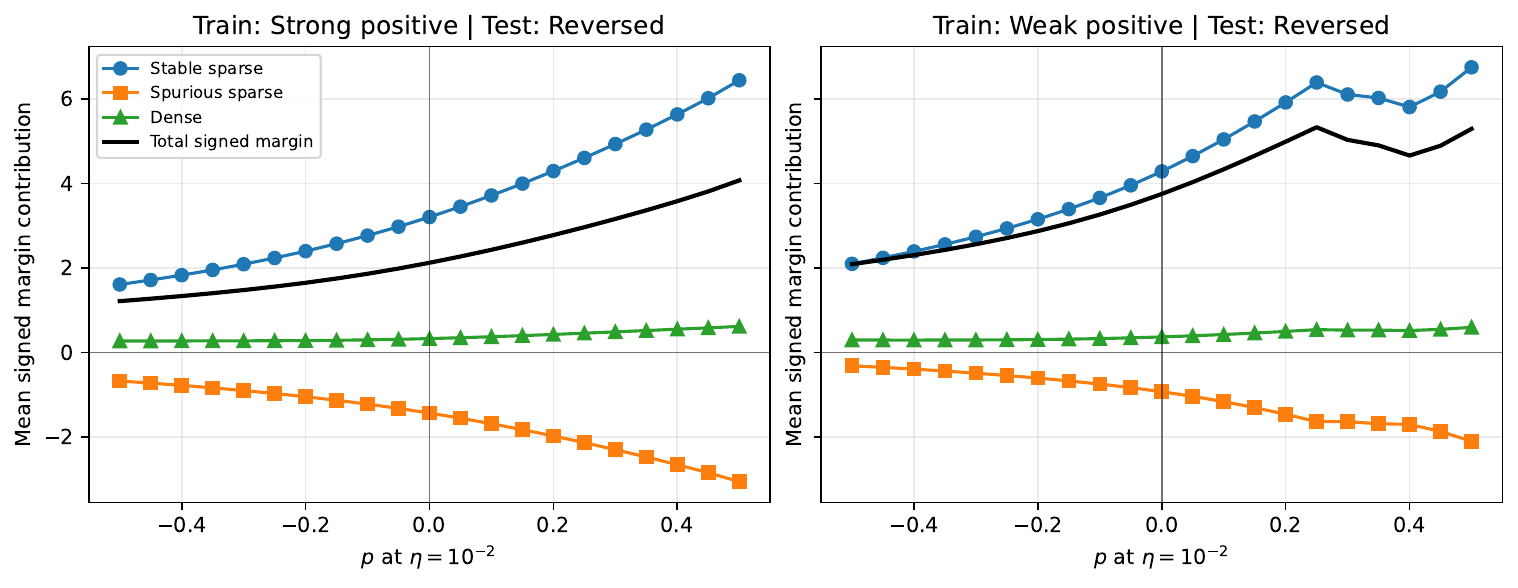}
  \caption{Mean signed margin decomposition on the reversed test environment for models trained on strong- and weak-positive sources at $\eta=10^{-2}$. The stable sparse contribution is positive and the spurious contribution is negative. Lower $p$ reduces the harmful spurious margin, explaining why the cross-environment optimum lies below the source-validation optimum.}
  \label{fig:margins}
\end{figure}

Weight norms do not by themselves show whether a feature block helps or harms a target environment. We therefore evaluate the signed margin contribution in \cref{eq:margin}. \Cref{fig:margins} considers models trained on positive-correlation environments and tested on the reversed environment. The stable sparse block contributes a positive margin, while the spurious block contributes a negative margin. As $p$ increases, the model learns more of both. The stable contribution grows, but so does the harmful reversed spurious contribution. The total margin is therefore maximized in a lower-$p$ region than source fitting alone would select.

This decomposition also clarifies why negative $p$ does not simply mean ``ignore sparse features.'' Both the stable and spurious blocks are sparse and use the same activation probabilities and signal magnitudes. Their difference is persistence across environments. The optimizer is not given this semantic distinction. Yet under the high-learning-rate trajectory, lower $p$ produces a solution with a smaller harmful spurious component relative to the stable component.

\subsection{Finite-budget character of the result}

In \FinalEpochFraction\% of the \NumRuns{} runs, the best source-validation checkpoint occurs at epoch 50, the end of the training budget. The experiment therefore primarily characterizes \emph{finite-budget feature allocation}, not the unique asymptotic solution of every configuration. This matters because changing $p$ can alter both the direction and speed with which different feature blocks are learned. A longer training horizon may narrow some differences or move the optimum. We retain this limitation explicitly because finite-budget training is also the practical regime in which optimizer choice matters.

\section{Discussion}

\subsection{Why can the robust optimum become negative?}

For ordinary Adam-like $p>0$, coordinates with small $\widehat v_i$ receive larger gains. In a high-dimensional sparse-noise setting, many weakly activated or inconsistent coordinates can have small second moments. A large global learning rate already supplies substantial step amplitude. Continuing to amplify low-$v$ coordinates can then over-allocate updates to weak, noisy, or source-specific directions. Decreasing $p$ counteracts this effect. Once $p<0$, low-$v$ coordinates are suppressed and coordinates with more sustained second-moment statistics receive relatively larger updates.

This interpretation is consistent with the checkpoint decomposition, but it is not a universal theorem that stable features always have larger $v$. The mapping depends on feature frequency, signal magnitude, label correlation, batch noise, and the evolving residual. The correct conclusion is conditional:

\begin{quote}
In this controlled sparse-feature regime, the learning-rate increase shifts the robust allocation toward lower $p$, and the shift crosses zero at sufficiently large $\eta$.
\end{quote}

\subsection{Learning rate and $p$ are not interchangeable}

The approximate compensation relation in \cref{eq:compensation} explains why the optimum follows a diagonal path in the grid. However, $\eta$ and $p$ remain distinct controls. The learning rate scales the entire current update and changes the next parameter state. The exponent changes coordinate ratios immediately through \cref{eq:coordinate_ratio}. Two configurations can have similar aggregate performance while assigning different weights to stable, spurious, and noise blocks. Therefore, $p$ cannot be reduced to a reparameterized learning rate.

\subsection{Optimizer model selection under distribution shift}

The source-validation optimum remains positive even when the robust optimum is negative. This finding creates a practical obstacle: target environments are unavailable in genuine domain generalization. The current experiment uses $p_{\mathrm{cross}}^\star$ and $p_{\mathrm{worst}}^\star$ only as diagnostic oracles. A deployable method would require a source-observable proxy for harmful allocation, perhaps based on gradient stability, moment concentration, feature perturbations, or multiple source splits. Developing such a criterion is outside the present study.

\subsection{Relation to partially adaptive and root-free methods}

Padam restricts $p$ to the interval between momentum-like and Adam-like updates and uses it to moderate adaptivity \citep{chen2020padam}. Root-free methods move in the opposite positive direction and can improve optimization and generalization in other architectures \citep{lin2024remove}. Our result does not contradict either line. It shows that the useful exponent is task-, learning-rate-, and criterion-dependent. The negative regime emerges here because the objective is cross-environment robustness under a large global step and a sparse high-dimensional feature competition.

\section{Limitations}

The main limitations are:
\begin{enumerate}[leftmargin=1.5em]
  \item \textbf{Single seed.} The paired construction removes substantial sampling variation, but it does not replace a multi-seed estimate of optimizer variability.
  \item \textbf{Linear model and synthetic data.} The feature blocks are directly interpretable, which is essential for mechanism analysis, but the result is not yet a claim about deep representations or natural datasets.
  \item \textbf{Finite training horizon.} Most checkpoints are selected at the final epoch. Additional long-horizon and matched-source-loss experiments are needed to separate transient allocation from asymptotic implicit bias.
  \item \textbf{Oracle robust selection.} Cross and worst-environment criteria use target test environments only for retrospective analysis. No source-only rule for selecting negative $p$ is proposed.
  \item \textbf{One optimizer family.} The experiment uses Adam-style first and second moments with a variable exponent. Momentum, moment decay, clipping, weight decay, and schedules may change the coupling.
\end{enumerate}

These limitations define a direct validation path: repeat a reduced grid across several seeds, extend representative runs to convergence, match source loss across configurations, and test whether source-observable allocation statistics predict the robust exponent.

\section{Conclusion}

A continuous preconditioning exponent and a global learning rate form a coupled optimization system. In a controlled four-environment sparse-feature problem, the exponent that maximizes cross-environment generalization decreases nearly linearly with the logarithm of the learning rate. At high learning rates, the robust optimum becomes negative even though source validation continues to prefer positive exponents. Negative $p$ reduces relative allocation to spurious and noise features and lowers the harmful spurious margin under correlation reversal. The main lesson is not that negative preconditioning is universally superior. It is that the sign and magnitude of the useful exponent depend on the step-size regime and on whether optimization is judged by source fit or environmental robustness.

\appendix

\section{Complete Optimal-Exponent Table}

\begin{table}[h]
  \centering
  \caption{Complete source-validation, mean-cross, and worst-environment optima for every learning rate. Accuracy values correspond to the criterion-specific robust optimum.}
  \label{tab:all-optima}
  \scriptsize
  \resizebox{\textwidth}{!}{\begin{tabular}{lrrrrrr}
\toprule
Source & $\eta$ & $p_{\mathrm{ID}}^\star$ & $p_{\mathrm{cross}}^\star$ & $p_{\mathrm{worst}}^\star$ & Cross acc. (\%) & Worst acc. (\%) \\
\midrule
Strong positive & 1e-04 & 0.50 & 0.45 & 0.45 & 88.79 & 78.22 \\
 & 3e-04 & 0.50 & 0.35 & 0.20 & 89.40 & 79.31 \\
 & 1e-03 & 0.50 & 0.20 & 0.05 & 89.90 & 80.08 \\
 & 3e-03 & 0.40 & 0.00 & -0.05 & 90.25 & 80.44 \\
 & 1e-02 & 0.25 & -0.05 & -0.10 & 90.43 & 80.63 \\
\addlinespace
Weak positive & 1e-04 & 0.50 & 0.40 & 0.30 & 91.69 & 84.19 \\
 & 3e-04 & 0.50 & 0.25 & 0.15 & 93.70 & 87.85 \\
 & 1e-03 & 0.40 & 0.10 & 0.10 & 95.18 & 90.61 \\
 & 3e-03 & 0.25 & 0.00 & -0.10 & 95.89 & 92.05 \\
 & 1e-02 & 0.15 & -0.20 & -0.30 & 96.39 & 93.16 \\
\addlinespace
Neutral & 1e-04 & 0.50 & 0.50 & 0.50 & 95.39 & 95.34 \\
 & 3e-04 & 0.50 & 0.35 & 0.30 & 96.61 & 96.52 \\
 & 1e-03 & 0.35 & 0.15 & 0.15 & 97.63 & 97.60 \\
 & 3e-03 & 0.25 & 0.05 & 0.05 & 97.99 & 97.95 \\
 & 1e-02 & 0.10 & -0.10 & -0.10 & 98.14 & 98.11 \\
\addlinespace
Reversed & 1e-04 & 0.50 & 0.25 & 0.15 & 83.44 & 75.70 \\
 & 3e-04 & 0.50 & 0.15 & 0.00 & 87.36 & 81.01 \\
 & 1e-03 & 0.35 & 0.00 & -0.15 & 90.05 & 85.06 \\
 & 3e-03 & 0.25 & -0.15 & -0.30 & 91.61 & 87.40 \\
 & 1e-02 & 0.10 & -0.30 & -0.45 & 92.54 & 88.90 \\
\addlinespace
\bottomrule
\end{tabular}}
\end{table}

\section{Source-Selected Cross-Environment Matrix}

For completeness, \cref{tab:source-matrix} reports test accuracy when both $p$ and $\eta$ are selected only by source validation loss. Rows are source environments and columns are test environments. This table is not the oracle robust result; it shows the performance of the deployable source-selection protocol used for checkpoint and hyperparameter selection in the original experiment.

\begin{table}[h]
\centering
\caption{Cross-environment accuracy (\%) after selecting $p$ and $\eta$ by source validation loss.}
\label{tab:source-matrix}
\begin{tabular}{lrrrr}
\toprule
Source & Strong pos. & Weak pos. & Neutral & Reversed \\
\midrule
Strong positive & 99.57 & 98.00 & 91.97 & 79.08 \\
Weak positive   & 99.51 & 98.50 & 95.58 & 90.93 \\
Neutral         & 97.79 & 97.79 & 97.77 & 97.84 \\
Reversed        & 84.97 & 91.26 & 96.17 & 98.62 \\
\bottomrule
\end{tabular}
\end{table}

\section{Reproducibility and Data Files}

The accompanying source package contains:
\begin{itemize}[leftmargin=1.5em]
  \item the original training script;
  \item the data-preparation and checkpoint-decomposition script;
  \item the complete manifest of 420 runs;
  \item all 1680 cross-environment evaluations;
  \item enriched run-level metrics with feature-block weight statistics and margin decompositions;
  \item compact CSV files used for each figure and table;
  \item vector PDF figures generated from those CSV files.
\end{itemize}
The exact experiment defaults are stored in \texttt{data/fixed\_config.json}. The main analysis files are \texttt{optimal\_p\_by\_learning\_rate.csv}, \texttt{optimal\_p\_linear\_fits.csv}, \texttt{high\_lr\_p\_curves.csv}, and \texttt{run\_metrics\_with\_feature\_allocation.csv}.

\bibliographystyle{plainnat}
\bibliography{references}

\begin{thebibliography}{12}
\providecommand{\natexlab}[1]{#1}
\providecommand{\url}[1]{\texttt{#1}}
\expandafter\ifx\csname urlstyle\endcsname\relax
  \providecommand{\doi}[1]{doi: #1}\else
  \providecommand{\doi}{doi: \begingroup \urlstyle{rm}\Url}\fi

\bibitem[Arjovsky et~al.(2019)Arjovsky, Bottou, Gulrajani, and
  Lopez-Paz]{arjovsky2019irm}
Martin Arjovsky, L{\'e}on Bottou, Ishaan Gulrajani, and David Lopez-Paz.
\newblock Invariant risk minimization.
\newblock \emph{arXiv preprint arXiv:1907.02893}, 2019.

\bibitem[Chen et~al.(2020)Chen, Zhou, Tang, Yang, Cao, and Gu]{chen2020padam}
Jinghui Chen, Dongruo Zhou, Yiqi Tang, Ziyan Yang, Yuan Cao, and Quanquan Gu.
\newblock Closing the generalization gap of adaptive gradient methods in
  training deep neural networks.
\newblock \emph{Proceedings of the Twenty-Ninth International Joint Conference
  on Artificial Intelligence}, pages 3267--3275, 2020.

\bibitem[Duchi et~al.(2011)Duchi, Hazan, and Singer]{duchi2011adagrad}
John Duchi, Elad Hazan, and Yoram Singer.
\newblock Adaptive subgradient methods for online learning and stochastic
  optimization.
\newblock In \emph{Journal of Machine Learning Research}, volume~12, pages
  2121--2159, 2011.

\bibitem[Gulrajani and Lopez-Paz(2021)]{gulrajani2021domainbed}
Ishaan Gulrajani and David Lopez-Paz.
\newblock In search of lost domain generalization.
\newblock In \emph{International Conference on Learning Representations}, 2021.

\bibitem[Kingma and Ba(2015)]{kingma2015adam}
Diederik~P. Kingma and Jimmy Ba.
\newblock Adam: A method for stochastic optimization.
\newblock \emph{International Conference on Learning Representations}, 2015.

\bibitem[Li et~al.(2019)Li, Wei, and Ma]{li2019largelr}
Yuanzhi Li, Colin Wei, and Tengyu Ma.
\newblock Towards explaining the regularization effect of initial large
  learning rate in training neural networks.
\newblock In \emph{Advances in Neural Information Processing Systems},
  volume~32, 2019.

\bibitem[Lin et~al.(2024)Lin, Dangel, Eschenhagen, Bae, Turner, and
  Makhzani]{lin2024remove}
Wu~Lin, Felix Dangel, Runa Eschenhagen, Juhan Bae, Richard~E. Turner, and
  Alireza Makhzani.
\newblock Can we remove the square-root in adaptive gradient methods? a
  second-order perspective.
\newblock In \emph{Proceedings of the 41st International Conference on Machine
  Learning}, volume 235, pages 29949--29973, 2024.

\bibitem[Lu et~al.(2024)Lu, Wu, Yang, and Zou]{lu2024benign}
Miao Lu, Beining Wu, Xiaodong Yang, and Difan Zou.
\newblock Benign oscillation of stochastic gradient descent with large learning
  rate.
\newblock In \emph{International Conference on Learning Representations}, 2024.

\bibitem[Qiu et~al.(2024)Qiu, Kuang, and Goel]{qiu2024complexity}
Guanwen Qiu, Da~Kuang, and Surbhi Goel.
\newblock Complexity matters: Feature learning in the presence of spurious
  correlations.
\newblock In \emph{Proceedings of the 41st International Conference on Machine
  Learning}, volume 235, pages 41658--41697, 2024.

\bibitem[Sagawa et~al.(2019)Sagawa, Koh, Hashimoto, and
  Liang]{sagawa2019groupdro}
Shiori Sagawa, Pang~Wei Koh, Tatsunori~B. Hashimoto, and Percy Liang.
\newblock Distributionally robust neural networks for group shifts: On the
  importance of regularization for worst-case generalization.
\newblock \emph{arXiv preprint arXiv:1911.08731}, 2019.

\bibitem[Wilson et~al.(2017)Wilson, Roelofs, Stern, Srebro, and
  Recht]{wilson2017marginal}
Ashia~C. Wilson, Rebecca Roelofs, Mitchell Stern, Nathan Srebro, and Benjamin
  Recht.
\newblock The marginal value of adaptive gradient methods in machine learning.
\newblock In \emph{Advances in Neural Information Processing Systems},
  volume~30, 2017.

\bibitem[Zhang et~al.(2026)Zhang, Liu, Ren, Sheng, Wang, and
  Liu]{zhang2026allocation}
Gongyue Zhang, Donghan Liu, Weihong Ren, Yixuan Sheng, Zhiyong Wang, and
  Honghai Liu.
\newblock Information allocation dynamics in neural network optimization.
\newblock \emph{arXiv preprint arXiv:2607.07156}, 2026.

\end{thebibliography}

\end{document}